**National and State-Level Approaches to Automated Vehicle Certification: A Comparative Framework for State Operational Governance**

**Sk Nahia Ahsan***
Graduate Research Assistant, Department of Civil, Construction & Environmental Engineering
North Carolina State University, Raleigh, NC 27606
Email: sahsan@ncsu.edu

**Md. Abdullah Al Hasan**
Graduate Research Assistant, Department of Civil, Construction & Environmental Engineering
North Carolina State University, Raleigh, NC 27606
Email: malhasa2@ncsu.edu

**Danjue Chen**
Associate Professor, Department of Civil, Construction & Environmental Engineering
North Carolina State University, Raleigh, NC 27606
Email: dchen33@ncsu.edu

**George List**
Professor, Department of Civil, Construction & Environmental Engineering
North Carolina State University, Raleigh, NC 27606
Email: gflist@ncsu.edu

*Corresponding Author

**ABSTRACT**

**Objectives:** Automated vehicle (AV) deployment in the United States is expanding despite the absence of a unified federal certification framework, resulting in substantial variation in how states authorize and oversee AV testing and deployment. This paper examines how U.S. states regulate Society of Automotive Engineers (SAE) Level 4–5 automated driving systems (ADS), identifies common certification approaches, compares them with international practices, and evaluates the implications for North Carolina's future regulatory strategy.

**Methods:** The study reviewed publicly available state statutes, administrative regulations, agency guidance documents, and federal policy documents governing AV deployment. Ten U.S. states, selected to capture the range of observed regulatory approaches, were analyzed using a structured framework covering operational scope, governance and oversight, federal interaction, and implementation outcomes. International certification practices from the European Union, China, Japan, and Australia were reviewed for comparative context, and findings were synthesized into a comparative framework of state certification models.

**Findings:** Three distinct state certification models were identified, representing increasing levels of regulatory involvement: baseline legal compliance, administrative self-certification with ongoing oversight, and permit-based authorization requiring formal agency review before deployment. North Carolina currently aligns with the baseline compliance model. International frameworks are converging toward scenario-based testing, operational-design-domain-specific approval, and safety-case documentation, cybersecurity assurance, and lifecycle oversight, although direct transferability to the U.S. is limited.

**Novelty:** This paper develops a structured three-model typology for comparing state AV certification approaches and integrates state-level regulatory practices with international certification frameworks, providing a consistent basis for evaluating differences in regulatory involvement, governance complexity, and institutional responsibilities across jurisdictions.

**Practical Applications:** The proposed framework offers transportation agencies and policymakers a practical decision-support tool for evaluating alternative approaches to AV governance, supporting future legislative and institutional planning by clarifying trade-offs among safety oversight, administrative burden, deployment flexibility, and organizational capacity.

## INTRODUCTION

The emergence of automated driving systems (ADS) at Society of Automotive Engineers (SAE) Levels 4 and 5 presents a fundamental challenge to existing vehicle safety and certification frameworks. Traditional regulatory pathways were designed for human-operated, mechanically oriented vehicles in which safety performance is assessed through static hardware tests and driver licensing. The integration of software-intensive systems, artificial intelligence (AI), sensor fusion, and continuous over-the-air updates introduces new dimensions of uncertainty that static one-time approval processes are ill equipped to address.

As ADS development progresses, driving responsibility increasingly shifts from the human driver to a computer-based system. In some configurations, the human is removed from the control loop entirely. This paradigm shift creates novel verification challenges across multiple fronts: AI-based perception and decision-making modules require specialized evaluation for model generalization and edge-case robustness; cybersecurity threats affect both safety-critical and non-safety-critical vehicle functions; and software updates alter system behavior over time in ways that periodic certification cannot track (Baldini 2020; Koopman and Wagner 2017).

The United States adds a structural complication: there is no unified federal regulatory framework governing ADS certification. The National Highway Traffic Safety Administration (NHTSA) exercises authority over vehicle design standards through a self-certification model under the Federal Motor Vehicle Safety Standards (FMVSS), but these standards do not explicitly address fully driverless systems. Operational permissions—testing, deployment, insurance, and enforcement—rest with individual states, producing significant regulatory fragmentation (Congressional Research Service 2025; Urban SDK 2025). In a 2025 survey, more than 93% of AV industry chief executives identified the establishment of a federal policy framework as the single most consequential action needed for safe deployment (AVIA 2025).

Against this backdrop, state governments serve as the primary operational governors of ADS on public roads. Understanding how states currently structure authorization and oversight—and how those structures compare internationally—provides essential context for states such as North Carolina that must decide how to calibrate their regulatory posture as AV technologies mature.

Although several studies and policy reports describe individual state regulations or federal guidance, a systematic framework that compares state operational governance models and synthesizes them into a unified classification remains lacking. Such a framework is needed to better understand differences in regulatory complexity, institutional responsibilities, and potential pathways for states considering future AV oversight.

This paper addresses the following research question: when an AV operator seeks to test or deploy an ADS within a U.S. state, how does the state regulate that activity, what requirements are imposed, and where do state practices converge or diverge? The analysis draws on publicly available state statutes, administrative rules, agency guidance documents, and interview-based insights from complementary NCDOT-sponsored research (Feng et al. 2026).

The remainder of the paper reviews the U.S. federal regulatory context, examines state-level certification practices and develops a three-model typology, compares these practices with international regulatory approaches, evaluates their implications for North Carolina, and concludes with key findings and considerations for future AV governance.

## FEDERAL REGULATORY CONTEXT

Federal authority over AV safety in the United States is exercised primarily through NHTSA under existing motor vehicle safety law. In the absence of AV-specific regulations, federal oversight relies on the traditional self-certification model: manufacturers certify that their vehicles comply with FMVSS, even though these standards were not designed with autonomous architectures in mind. NHTSA has supplemented this framework with non-binding policy guidance—Automated Driving Systems 2.0, 3.0, and 4.0—covering system design, operational domains, and cybersecurity considerations, but these documents do not constitute regulatory approval or a formal certification process (NHTSA 2017, 2020, 2021).

Post-deployment federal oversight operates through defect investigations, mandatory crash reporting for ADS-equipped vehicles under a Standing General Order, and recalls. Limited FMVSS exemptions are available to facilitate testing, but they are constrained in scope and unsuitable for large-scale commercial deployment. Key issues—liability assignment, cybersecurity standards, and operational governance—remain largely unresolved at the federal level and have been addressed piecemeal by states (Congressional Research Service 2025).

Potential future directions under discussion include minimum standards for ADS safety behaviors (e.g., fallback performance, object detection benchmarks), operational design domain (ODD)-specific certification approvals, nationally standardized scenario-based performance tests, safety-case submission requirements aligned with UL 4600 and ISO 21448, and continuous post-market data reporting requirements (NHTSA 2017, 2021; RAND Corporation 2018; Congressional Research Service 2025).

The current absence of a unified national framework has led to a patchwork of state permitting systems and substantial variation in how ADS operations are authorized and monitored across the country.

## STATE LEVEL CERTIFICATION PRACTICES

### Analytical Framework

Ten U.S. states — California, the District of Columbia, Pennsylvania, Texas, Washington, Arizona, Georgia, North Carolina, Wyoming, and Michigan — were selected for review to capture the range of observed regulatory approaches, from minimal baseline compliance to formal permit-based authorization, and to reflect states with documented AV testing or deployment activity. To support a systematic review, a set of analytical questions covering four dimensions was developed: (A) scope and operational conditions, (B) governance, evidence requirements, and oversight, (C) interaction with federal policy and cross-state portability, and (D) implementation outcomes as shown in **Table 1**. These questions were applied consistently across all reviewed jurisdictions to enable structured comparison and identification of common regulatory patterns. Sources included state statutes, administrative rules, agency guidance documents, official websites, and interviews conducted in January–February 2026.

The synthesis focuses on SAE Level 4–5 vehicles—commonly described in state statutes as 'fully autonomous vehicles'—because Level 4 ADS increasingly enables driverless operation and therefore attracts more stringent and consequential state oversight. Level 3 deployments remain limited in practice.

**TABLE 1 Questions Used to Review State AV Certification Practices**

| Category | Question ID | Question |
|---|---|---|
| A. Scope and Operational Conditions | A1 | What certification or authorization model is used (e.g., self-certification, submission-based approval, authority-issued permit)? |
| | A2 | Does certification or authorization differentiate by AV vehicle type or use case? |
| | A3 | Is certification or authorization tied to specific operational phases (e.g., testing, pilot, deployment)? |
| | A4 | Is certification or authorization explicitly conditioned on an Operational Design Domain (ODD)? |
| B. Governance, Evidence, and Oversight | B1 | Which entity is responsible for issuing AV certification or authorization? |
| | B2 | How are roles and responsibilities allocated between state and local governments? |
| | B3 | What types of evidence or documentation are typically required for certification or authorization? |
| | B4 | Under what conditions is re-certification or renewal required? |
| | B5 | Under what conditions can certification or authorization be suspended or revoked? |
| | B6 | What reporting obligations apply after deployment? |
| C. Federal Interaction and Portability | C1 | Is AV certification or authorization portable or recognized across states? |
| | C2 | How does state certification or authorization interact with federal policy and requirements? |
| D. Implementation Outcomes | D1 | Which AV developers or operators have obtained certification or authorization? |

**A Three-Model Typology**

Comparative analysis of the reviewed jurisdictions identified a consistent pattern that can be generalized into three regulatory models representing increasing levels of regulatory involvement and administrative workload. Higher tiers generally incorporate the requirements of lower tiers.

- Tier 1 — Baseline Legal Compliance: Requirements applicable universally, including self-certified compliance with applicable federal standards and state traffic laws, vehicle registration as a fully autonomous vehicle, and state-minimum liability insurance. Compliance is implicit and grounded in existing law without AV-specific administrative programs.
- Tier 2 — Administrative Oversight and Self-Certification: Explicit AV-specific administrative elements including safety management plan submission (notably law enforcement interaction procedures), self-certification through an authority-designated platform, and post-deployment reporting obligations such as crash reporting.
- Tier 3 — Authority-Issued Permit or Authorization: An explicit permit issued by a public authority prior to any AV testing or operation, entailing a formal review process, defined application requirements, and structured conditions for modification, renewal, and revocation.

Based on these tiers, the reviewed jurisdictions naturally cluster into three regulatory groups. **Table 2** summarizes the relationship between individual requirements and the resulting classification.

**TABLE 2 State Certification Models**

| **Requirement Item** | | **Regulatory Workload & Capacity Needs** | **Group 1 GA, NC** | **Group 2 AZ, WA** | **Group 3 CA, PA, TX, DC** |
|---|---|---|---|---|---|
| **Tier 1** | Compliance with federal standards & state laws | Light | Y | Y | Y |
| | Registration | | Y | Y | Y |
| | Insurance | | Y | Y | Y |
| **Tier 2** | Law enforcement coordination plan | Moderate | No | Y | Y |
| | Self-certify via authority channel | | No | Y | Y |
| | Reporting | | No | Y | Y |
| **Tier 3** | Permit prior to operate | High | No | No | Y |

Group 1 — Minimalist (e.g., Georgia and North Carolina): These states use Tier 1 requirements only. AV operation is permitted once baseline legal conditions are met, without AV-specific administrative programs or permitting frameworks. Authorization is largely implicit, grounded in compliance with existing federal and state laws.

Group 2 — Administrative Self-Certification (e.g., Arizona and Washington): These states incorporate Tier 2 requirements in addition to Tier 1. Formal administrative processes are established— safety plan submission, self-certification through official platforms, and reporting obligations—without requiring an authority-issued permit prior to operation. Authorization is based on documented compliance and post-deployment oversight.

Group 3 — Permit-Based Authorization (e.g., California, Pennsylvania, Texas, and the District of Columbia): These states require an explicit permit or authorization prior to AV testing or operation, through a formal application and review process. Permitting frameworks generally impose more elaborate Tier 2 requirements and may expand to include renewal conditions, operational constraints, data retention requirements, and differentiated permit types by phase or vehicle class.

It is worth noting that within Group 3, there is substantial variation in both the level of program elaboration and the stringency of permitting requirements. Although all Group 3 states require authority-issued authorization prior to AV testing or operation, the scope, detail, and procedural rigor of these permitting frameworks differ significantly.

California operates the most elaborate and prescriptive program among Group 3 states. Its framework differentiates permits by operational phase (e.g., testing with a safety driver, driverless testing, deployment) and by vehicle type (e.g., seating capacity), and includes detailed reporting obligations and clearly defined renewal and modification conditions. The program reflects a relatively high level of regulatory stringency and well-defined procedures. The program is supported by a distinct regulatory structure involving two primary agencies: the Department of Motor Vehicles (DMV), which handles the technology and vehicle aspects to issue the prerequisite permits, and the California Public Utilities Commission (CPUC), which regulates the commercial service of carrying passengers. Permits from both agencies are required for passenger-carrying operations.

Pennsylvania also provides detailed certification guidance; however, much of this material is framed as guidance rather than mandates. As a result, while expectations for AV operation are articulated with substantial details, formal permitting requirements are less prescriptive than in California. Overall, the program is less stringent than California's.

Texas requires authorization for AV operation but maintains comparatively brief statutory requirements. The authorization framework establishes baseline conditions (e.g., presence of a recording device) while avoiding detailed procedural or reporting specifications, resulting in a lower level of elaboration within the permit-based model. Texas only recently adopted a formal permit requirement (effective September 1, 2025), corresponding with increased AV deployment activity and emerging safety concerns (Texas Department of Motor Vehicles [TxDMV] 2025; Robles 2025).

The District of Columbia applies a highly restrictive permitting approach, currently requiring a human safety driver behind the wheel for AV operation. The existing permitting framework is interim. A formal framework addressing operation both with and without safety drivers is anticipated and is likely to involve elaborate requirements and a formalized approval process. The district's permit requirement is also relatively recent, with emergency legislation first enacted in March 2024 (District of Columbia Council 2024).

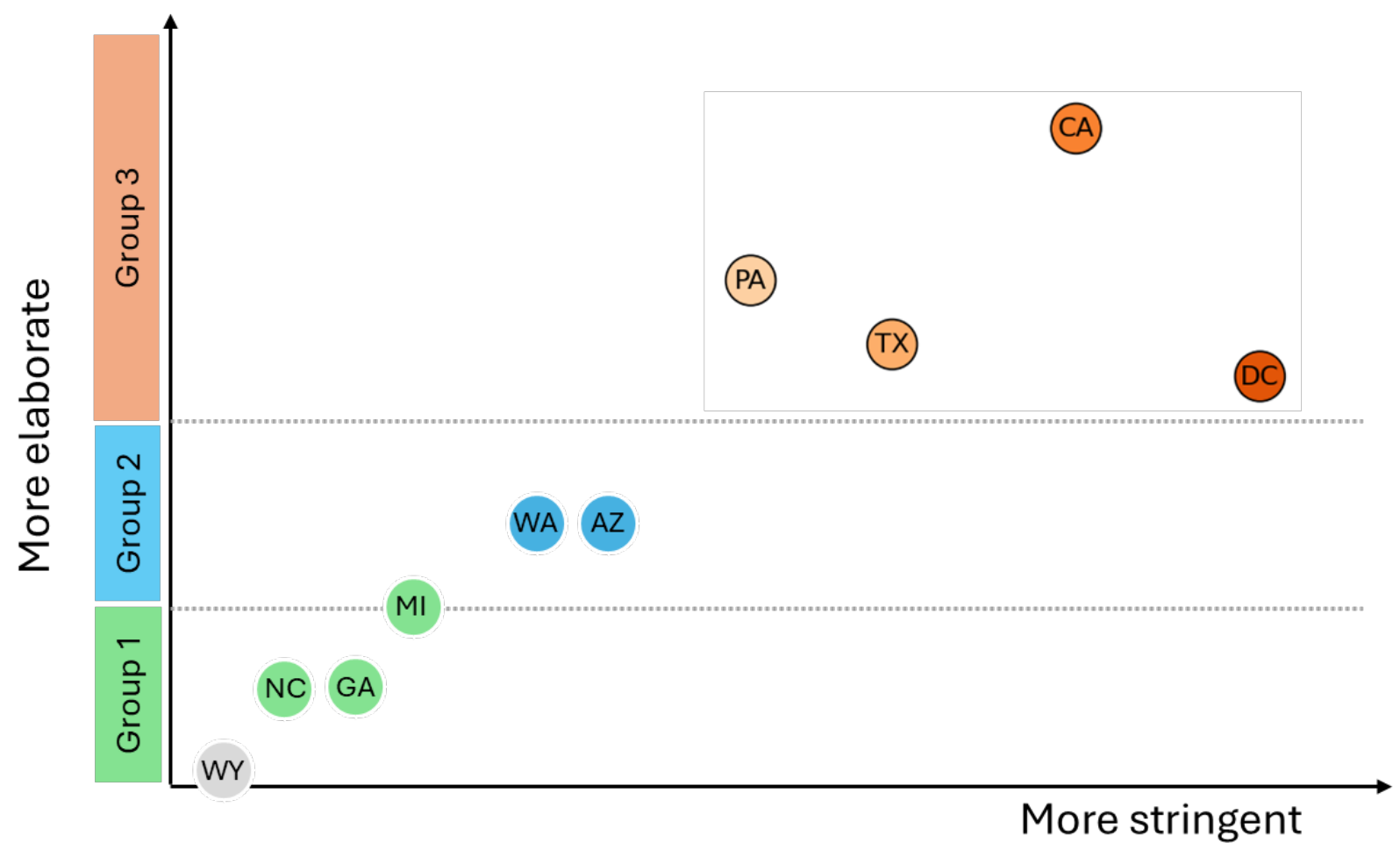

**Figure 1 Conceptual positioning of reviewed states according to relative regulatory elaboration and stringency**

The states have been positioned across the three groups in **Figure 1** using a 2-dimensional plot to illustrate relative levels of elaboration and stringency.

While most states align with one of the three groups, several cases fall between categories. For example, Michigan requires that prior to research testing, manufacturer or upfitter must be recognized and submit proof of insurance to the Secretary of State (Mich. Comp. Laws § 257.665 et seq. 2023). However, Michigan has removed permit systems to remain flexible. Thus, Michigan's model shows features spanning Group 1 and 2. Wyoming represents another situation in which AV operations remain unregulated. In Wyoming, AV operators are encouraged, but not required, to notify the Wyoming Department of Transportation and Wyoming Highway Patrol prior to entry. Please note that these positions only reflect the status as of January 2026 and they may change as legislation evolves.

Notably, the different models entail distinct levels of workload and institutional capacity requirements for regulating agencies.

- Group 1 model: Light workload and minimum institutional changes. Implementation is typically embedded within existing regulatory frameworks rather than AV-specific programs, requiring little modification to current institutional arrangements.
- Group 2 model: Moderate workload and institutional complexity. This model introduces Tier 2 requirements, which require regulatory authorities to review submitted documentation (e.g., safety management plans), conduct post-deployment oversight, and exercise regulatory judgment (e.g., suspending or restricting operations when warranted). As a result, additional institutional considerations arise, including which agencies are best positioned to perform these functions and how responsibilities should be allocated. Multiple agencies may be involved as in Arizona, where both the State Department of Transportation and Department of Public Safety are involved. When multiple agencies are designated, inter-agency coordination becomes an important consideration.
- Group 3 model: High workload and institutional design complexity. This model entails formal permitting processes, pre-operation review, and sustained regulatory oversight, placing the greatest demands on administrative capacity and institutional design.

Detailed certification practices for all reviewed states are summarized in **Table 3, Table 4, Table 5.**

**TABLE 3 AV Certification Practice — California, District of Columbia, Pennsylvania, and Texas**

| Category | Question | California (CA) | District of Columbia (DC) | Pennsylvania (PA) | Texas (TX) |
|---|---|---|---|---|---|
| **A. Scope & Operational Conditions** | **A1. Certification model used?** | Permit-based; formal tiered permit program. | Interim permit program; moving toward formal permitting. | Self-certification with elaborate guidelines. | Authorization-based; state permit required for driverless AVs. |
| | **A2. Specific to vehicle type or use case?** | Yes — differentiates by operational phase and seating capacity. | Distinguishes testing vs. deployment; requires in-vehicle safety operators. Focus on L3+. | Applies to Highly Automated Vehicles (HAVs), SAE Levels 3–5. | Driverless AVs only (SAE L4–L5). |
| | **A3. Tied to specific operational phases?** | Yes — Driver Pilot, Driverless Pilot, Phase I Driver Deployment, Phase I Driverless Deployment. | Testing allowed; deployment legislation still developing. | Approval issued in perpetuity unless material change occurs. | Authorization required prior to driverless public-road operation. |
| | **A4. Conditioned on ODD?** | Yes — authorization is contingent on the certified ODD. | Uses geofences; retains authority to order vehicles off road for unique events. | ODD defined by developer; expansion requires re-application. | No state-defined ODD. |
| **B. Governance, Evidence & Oversight** | **B1. Which entity issues approval?** | Two agencies: DMV (technology/vehicle) and CPUC (commercial passenger service). Both required. | Department of Transportation (oversight moved from DMV). | PennDOT. | Texas Department of Motor Vehicles (TxDMV). |
| | **B2. Role of state vs. local government?** | State agencies primary; local agencies have very limited authority. | City-state; direct control via Council/Mayor. | PennDOT is sole authority; municipalities prohibited from enacting own regulations. | State maintains authorization authority; local governments explicitly prohibited from regulating driverless vehicles (SB 2205). |
| | **B3. Evidence typically required?** | Varies by permit type; deployment permit requires: DMV AV Manufacturer's Permit, prior 30-day CA road | Certified driving records for remote operators; training frameworks. | Safety Management Plan (driver training, response procedures), insurance. | Authorization materials: safety documentation, insurance, recording device, emergency |
| | | testing, TCP (California charter-party permit) authority, ODD documentation, law enforcement & collision plan, passenger safety plan. | | | interaction info, FMVSS compliance. |
| | **B4. When is re-certification required?** | TCP Permit Renewal every 3 years; material changes (ODD updates, added shared rides). | Not mentioned. | Material changes (ODD expansion, vehicle type change, driverless transition). | No periodic re-certification; updates required within 30 days of material changes. |
| | **B5. When can it be suspended or revoked?** | Administrative failures or extended regulatory review. | Implicit authority to restrict road access for safety/security. | If company operates outside conditions of self-certification. | Failure to update required info; failure to respond to TxDMV; unsafe operation endangering the public (with notice and opportunity to cure). |
| | **B6. Reporting requirements?** | Extensive; CPUC requires all DMV reports (collision, annual disengagement), quarterly reports, and operational/financial reporting. | (Proposed) digital data via Mobility Data Specification for geofence monitoring. | Crash reporting within 6 hours. | Standard crash reporting; additional reporting may be imposed via authorization conditions. |
| **C. Federal Interaction & Portability** | **C1. Certification portable across states?** | Not addressed. | Not mentioned. | Not mentioned; PennDOT has authority to negotiate interstate agreements. | Not addressed. |
| | **C2. Interaction with federal policy?** | FMVSS and NHTSA rules baseline. | National security concerns involving FBI and Homeland Security. | Uses NHTSA crash data for context. | FMVSS and NHTSA definitions remain baseline. |
| **D. Implementation Outcomes** | **D1. Certified / authorized companies (examples)?** | Waymo, Zoox, Tensor Auto, Apollo Autonomous Driving USA, WeRide Corp. | Waymo, Zoox. | Waymo, Zoox, May Mobility. | Not publicly listed. |

**TABLE 4 AV Certification Practice — Washington, Arizona, Georgia, and North Carolina**

| Category | Question | Washington (WA) | Arizona (AZ) | Georgia (GA) | North Carolina (NC) |
|---|---|---|---|---|---|
| **A. Scope & Operational Conditions** | **A1. Certification model used?** | Self-certification via designated channels. | Submission-based via designated channels. | Self-certification. | Self-certification. |
| | **A2. Specific to vehicle type or use case?** | Applies to ADS-equipped vehicles (SAE L4–L5) tested or operated on public roads. | No; requirements specified for fully autonomous vehicles (SAE L4–L5). | No specific requirements by vehicle type or use case. | Applies to ADS-equipped vehicles. |
| | **A3. Tied to specific operational phases?** | Registration required prior to any L4+ testing or operation. | No. | No. | No. |
| | **A4. Conditioned on ODD?** | No state-level ODD; self-certification addresses ADS performance within its operational limits and compliance with traffic law. | Implicit only; operator self-attests safe operation and ability to reach a minimal risk condition; no ODD review or approval. | No; ODD defined by developer. | No state-defined ODD. |
| **B. Governance, Evidence & Oversight** | **B1. Which entity issues approval?** | Department of Licensing (DOL), designated as the exclusive state agency regulating AV operation. | No approval issued. ADOT (primary regulator) manages submissions and may suspend or restrict operations; DPS maintains the Law Enforcement Interaction Protocol and reviews interaction plans. | Dept. of Revenue (registration as a fully autonomous vehicle); Dept. of Public Safety (enforcement). | No authority-issued permit; registration presumably handled by NCDMV (not explicitly specified). |
| | **B2. Role of state vs. local government?** | DOL is exclusive; other state agencies and local entities are prohibited from regulating or limiting AV operation. | ADOT and DPS are the only state agencies that may implement or enforce rules; local governments prohibited from imposing additional | GDOT reviews concept plans on request and encourages local coordination; state law prohibits new rules further limiting fully | Local governments prohibited from enacting laws related to AV operation. |
| | | | taxes, fees, or requirements. | autonomous vehicle operation. | |
| | **B3. Evidence typically required?** | Law enforcement interaction plan to State Patrol; proof of financial responsibility and most recent NHTSA voluntary self-assessment to DOL; notice to local law enforcement within 14 days of operation (until 12/31/2028); AV registration and title. | Law enforcement interaction plan to ADOT and DPS (consistent with the DPS protocol of May 2018); written statement to ADOT affirming federal compliance, minimal risk condition capability, traffic law compliance, and title, registration, licensing, and insurance. | Manufacturer self-certification (FMVSS and state law compliance, post-crash duties, minimal risk condition); registration as an AV (in GA or another state); liability insurance or self-insurance per Code Section 40-1-166. | Manufacturer self-certification (FMVSS and state law compliance, post-crash duties, minimal risk condition); registration as an AV in NC; liability insurance per G.S. 20-279.21. |
| | **B4. When is re-certification required?** | Not specified; annual collision and moving violation report due each February 1. | Certification of correction or adjustment required if ADOT finds a vehicle is likely not in safe mechanical condition. | N/A. | Not mentioned. |
| | **B5. When can it be suspended or revoked?** | Not specified. | Registration suspended or restricted if a certification of correction is not submitted or is inaccurate; cease-and-desist if the interaction plan is not submitted. | N/A. | Not mentioned. |
| | **B6. Reporting requirements?** | Annual AV Collision Report (collisions and moving violations) to DOL and municipalities until 2028; on-scene accident reporting remains mandatory. | Crashes must be promptly reported to law enforcement with owner and registration information. | No. | No. |
| **C. Federal Interaction & Portability** | **C1. Certification portable across states?** | Not mentioned. | Not applicable (no certification). | Yes; AVs lawfully registered in other states may operate in GA. | Not mentioned. |
| | **C2. Interaction with federal policy?** | FMVSS compliance required, including reference to any NHTSA exemption. | Must comply with FMVSS; certification labels reference any NHTSA exemptions. | FMVSS. | FMVSS and NHTSA definitions remain baseline. |
| **D. Implementation Outcomes** | **D1. Certified / authorized companies (examples)?** | Waymo, Zoox, NVIDIA (listed by DOL). | Waymo, Aurora, May Mobility, Beep. | Not publicly listed. | Not publicly listed. |

**TABLE 5 AV Certification Practice — Wyoming and Michigan**

| Category | Question | Wyoming (WY) | Michigan (MI) |
|---|---|---|---|
| **A. Scope & Operational Conditions** | **A1. Certification model used?** | Voluntary notification only; no state-led certification or permitting. | No permit system; recognition and proof of insurance required by Secretary of State. |
| | **A2. Specific to vehicle type or use case?** | N/A — no certification requirement. | Yes — applies to manufacturers and upfitters of ADS; also covers SAVE projects (on-demand automated networks) and participating fleets. |
| | **A3. Tied to specific operational phases?** | N/A. | Applies for research or testing; also for SAVE project operations. |
| | **A4. Conditioned on ODD?** | No; vehicles must comply with standard road closures (e.g., wind/snow). | Implicit only; operator self-attests safe operation and ability to reach minimal risk condition; no ODD review or approval. |
| **B. Governance, Evidence & Oversight** | **B1. Which entity issues approval?** | N/A — no approval issued. | Secretary of State responsible for recognizing manufacturers and upfitters of ADS and receiving proof of insurance. |
| | **B2. Role of state vs. local government?** | Voluntary state notification; local governments may draft ordinances. | Undefined in statute. |
| | **B3. Evidence typically required?** | AV operators are encouraged (not required) to notify state DOT and Highway Patrol prior to entry. | Prior to research or testing: manufacturer or upfitter must be recognized by Secretary of State and submit proof of insurance under Chapter 31 of the Insurance Code of 1956. |
| | **B4. When is re-certification required?** | N/A. | N/A. |
| | **B5. When can it be suspended or revoked?** | N/A. | N/A. |
| | **B6. Reporting requirements?** | No. | N/A; encourages close operating discussion with DOT and law enforcement. |
| **C. Federal Interaction & Portability** | **C1. Certification portable across states?** | N/A. | Not addressed. |
| | **C2. Interaction with federal policy?** | Relies on guidance from NHTSA and AAMVA. | Non-automakers required to pass FMVSS. |
| **D. Implementation Outcomes** | **D1. Certified / authorized companies (examples)?** | Not publicly listed. | Not publicly listed. |

## GLOBAL PERSPECTIVE

### Overview

Viewing U.S. practices against international models illuminates both the distinctive features of the U.S. decentralized approach and the direction of global regulatory convergence. **Table 6** provides a structured overview of how major regions approach AV certification.

**TABLE 6 Perspectives on AV Certification Around the Globe**

| Region | Main Approach | Style | Key Strength | Key Weakness |
|---|---|---|---|---|
| U.S. | Self-certification + state permits | Emerging hybrid (not yet defined) | Innovation-friendly, flexible | No unified certification test; fragmented oversight |
| European Union | UNECE type approval | Formal, test-based, scenario-driven | Most mature and coherent global model | Slower regulatory cycles; limited ODD flexibility |
| China | National standards + mandatory scenario testing | Centralized and mandatory | Fast-moving; extensive scenario validation at scale | Opaque decision-making; IP/data access issues; city fragmentation |
| Japan | ODD-based approval + safety case | Cautious, safety-focused | World's first L3 certified vehicle; structured safety cases | Slow, individualized; scalability concerns |
| Australia | Proposed national safety law | Safety-case, performance-based | Clean-slate design; balanced safety case and metrics | Not yet implemented |

### United States: An Emerging Hybrid Framework

As noted in Section 2, the U.S. regulatory philosophy rests on self-certification and state-level operational permitting, without a unified federal certification program. NHTSA exercises oversight through recalls, investigations, and voluntary guidance. Strengths include rapid innovation and extensive test deployment by developers such as Waymo, Cruise, Aurora, and Kodiak through state testing permit programs. Limitations include uneven safety oversight, no national-level validation tests, and public trust concerns following major AV crash events (NTSB 2022; RAND Corporation 2018; Litman 2025).

### European Union: UNECE Type Approval with Scenario Testing

The EU employs a unified type-approval framework anchored in UNECE WP.29, supported by UN Regulation No. 157 (Automated Lane Keeping Systems), UN Regulation No. 155 (Cybersecurity), and UN Regulation No. 156 (Software Updates). Testing requirements include defined scenario suites for lane keeping, obstacle avoidance, emergency braking, and vulnerable road user protection, combining on-road and closed-course validation with simulation evidence. The framework provides the most complete and harmonized international certification system, though regulatory update cycles are slower and ODD flexibility remains limited (European Commission 2022; UNECE 2021a, 2021b).

### China: Centralized Standards with Mandatory Scenario Verification

China adopts a centralized approach with strong national control and rapidly evolving technical standards from MIIT and MPS (China's regulators), implemented through designated pilot zones in Beijing, Shanghai, and Shenzhen. Mandatory elements include scenario-based testing libraries, closed-course and public-road validation, cybersecurity reviews, data localization requirements, and real-time data access for regulators. Large-scale real-world AV fleets operated by Baidu Apollo and Pony.ai demonstrate the rapid deployment pace. Limitations include opacity for external stakeholders and heavy data compliance burdens (CAICT 2024).

**Japan: ODD-Based Approval and Safety Case**

Japan requires formal government approval for each ADS with strong emphasis on ODD documentation. Honda's 2021 approval for the Level 3 Sensing Elite system—the world's first commercially available L3 consumer vehicle—required safety-case documentation, hazard analysis, ODD mapping, redundant hardware validation, and human-machine interface assessment. Japan is moving toward Level 4 robotaxi operations in limited zones during 2025–2030, with close collaboration between national government and major OEMs including Honda, Nissan, and Toyota (METI 2023; Law.asia 2024).

**Australia and Other Regions**

Australia is designing a centralized national certification framework built around a mandatory safety case, risk assessment, deploy-and-monitor program, incident reporting system, and change-management system, though implementation is not yet complete. Singapore pursues safety-case approval for each ODD with mandatory simulation-based scenario testing and strong emphasis on remote operator requirements. South Korea combines type-approval with a pilot zone system and cybersecurity integration under its 2020 National AV Act. Dubai has the fastest regulatory approval process outside China, involving mandatory third-party certification and a focus on fleet-based AV services.

**Comparative Summary of Global Approaches**

**Table 7** provides structured comparisons across the five major regions on certification components, software integrity, performance recording and data sharing, review and oversight elements, and reporting requirements.

**TABLE 7 Comparison of Global AV Certification Approaches**

| **Overview of Regulatory Strategies Across Major Regions** | | | | | |
|---|---|---|---|---|---|
| **Category** | **United States** | **European Union** | **China** | **Japan** | **Australia** |
| Regulatory philosophy | Industry-led self-certification | Centralized and harmonized | Centralized national control | Cautious, ODD-based | Safety-case, performance-based |
| Main authority | NHTSA + state DOTs | UNECE WP.29 / national ministries | MIIT, MPS, local pilot zones | MLIT (Transport ministry) | National Transport Commission |
| Federal/unified rules | ✗ | ✔ | ✔ | ✔ | ✔ proposed |
| Speed of regulation | Slow | Medium | Very fast | Slow–medium | Medium |
| **Certification Components** | | | | | |
| **Category** | **U.S.** | **EU** | **China** | **Japan** | **Australia** |
| Type approval required | ✗ | ✔ | ● city/national | ✔ | ● proposed |
| Safety case required | ● voluntary (UL 4600) | ✔ integrated | ✔ required | ✔ required | ✔ required |
| Scenario-based testing | ● developing | ✔ mandatory | ✔ mandatory | ✔ limited | ✔ proposed |
| ODD-specific certification | ● developing | ✔ emerging | ✔ pilot zones | ✔ core | ✔ proposed |
| Simulation evidence required | ● informal | ✔ mandatory | ✔ mandatory | ✔ used | ✔ proposed |
| Closed-course testing | • | ✔ | ✔ | ✔ | ✔ |
| National scenario library | ✖ | ✔ in development | ✔ | • | • |
| **Software Integrity** | | | | | |
| **Category** | **U.S.** | **EU** | **China** | **Japan** | **Australia** |
| Functional safety (ISO 26262) | ✔ industry-led | ✔ required | ✔ required | ✔ required | ✔ required |
| SOTIF (ISO 21448) integration | • | ✔ | ✔ | ✔ | ✔ |
| Cybersecurity certification | • developing | ✔ UNECE R155 | ✔ mandatory | ✔ required | ✔ proposed |
| Software update certification | ✖ | ✔ UNECE R156 | ✔ required | ✔ required | • |
| HMI and fallback requirements | • | ✔ | ✔ | ✔ | ✔ |
| **Performance Recording and Data Sharing** | | | | | |
| **Category** | **U.S.** | **EU** | **China** | **Japan** | **Australia** |
| Robotaxi deployment approvals | State and local | National | City and national | National | Proposed |
| Data logging requirements | • | ✔ | ✔ | ✔ | ✔ |
| Real-time data access to regulators | ✖ | ✖ | ✔ required | ✖ | ✖ |
| Remote operator certification | ✖ | ✔ emerging | ✔ required | ✔ | ✔ |
| Geographic ODD fencing | • | ✔ | ✔ | ✔ | ✔ |
| **Review and Oversight Elements** | | | | | |
| **Category** | **U.S.** | **EU** | **China** | **Japan** | **Australia** |
| Recalls for AV systems | ✔ NHTSA | ✔ | ✔ | ✔ | ✔ |
| Mandatory incident reporting | • varies by state | ✔ | ✔ | ✔ | ✔ |
| Continuous oversight | • | ✔ | ✔ | • | ✔ planned |
| National AV registry | ✖ | ✔ | ✔ | ✖ | ✔ planned |
| **Reporting Requirements** | | | | | |
| **Category** | **U.S.** | **EU** | **China** | **Japan** | **Australia** |
| Public reporting of disengagements | • CA only | ✖ | ✖ | ✖ | ✖ |
| Required safety reports | ✖ voluntary | ✔ | ✔ | ✔ | ✔ |
| Required crash reports | ✔ | ✔ | ✔ | ✔ | ✔ planned |
| Testing database access | ● NHTSA AV TEST | ✔ | ✔ | ● | ● |

*Note:* ✔ *= Yes / Mature;* ● *= Partial / Emerging;* ✘ *= No / Absent*

Across the globe, there is convergence toward scenario-based testing (ISO 34502), ODD-specific certification, safety-case requirements (ISO 21448 / UL 4600), cybersecurity certification (ISO/SAE 21434; UNECE R155), continuous post-market monitoring, and digital infrastructure integration. Europe leads in regulatory completeness and harmonization; China leads in large-scale scenario testing and deployment data; Japan maintains a structured safety-focused program; Australia is designing an integrated national model. The United States focuses on technology development but lags in unified certification (NTSB 2022; RAND Corporation 2018).

## IMPLICATIONS FOR NORTH CAROLINA

### Current Position

North Carolina currently aligns with the Group 1 baseline legal compliance model, similar to Georgia. AV operation is authorized through compliance with existing federal and state laws—including FMVSS self-certification, registration as a fully autonomous vehicle, and minimum liability insurance under G.S. 20-279.21—without AV-specific administrative programs or permitting frameworks. Local governments are prohibited from enacting additional AV-related regulations. This approach minimizes administrative burden and preserves deployment flexibility but provides limited proactive oversight as driverless operations expand.

### The State Role: Operational Governance

Even absent formal technical certification authority, state agencies occupy a central position in AV governance. Federal authorities retain responsibility over vehicle design standards, recalls, and post-market enforcement. State departments of transportation and motor vehicles sit at the frontline of operational governance: testing authorization, operational rule-setting, vehicle registration, enforcement, and coordination with law enforcement and emergency responders (NHTSA 2017; NCSL 2020; AAMVA 2018).

North Carolina may also contribute to federal and national frameworks by raising state-specific test cases, advocating for inclusion of state operating conditions in federal standards, and seeking transparency and advisory participation in certification and oversight processes as they develop (RAND Corporation 2018; NTSB 2022).

### Pathways Toward More Elaborate Oversight

A shift from Group 1 toward Group 2 or Group 3 would expand the scope of operational governance without introducing technical certification responsibilities—algorithm validation, software testing, and hardware certification would remain with manufacturers or federal authorities. If North Carolina were to pursue more elaborate oversight, state-designated authorities could be expected to undertake the following functions:

- Establishing operational rules governing permitted vehicle classes, SAE automation levels, geographies, roadway types, speed limits, and operating conditions including ODD boundaries.
- Developing administrative mechanisms for authorization such as enhanced registration requirements, expanded insurance provisions, safety management plan submissions, law enforcement interaction protocols, emergency response procedures, and post-deployment reporting obligations. For driverless operations specifically, requirements related to remote operators would be critical, including operator-to-vehicle ratios and workload limits.
- Updating incident response capabilities, including AV-informed crash investigation protocols, guidance for law enforcement interactions with driverless vehicles, and procedures for accessing AV operational data to support accountability and liability assessment.

- Creating public information mechanisms, including data-sharing protocols within and across state agencies and public-facing transparency tools such as AV operational dashboards.

A transition toward more elaborate oversight would likely require legislative action and would involve balancing potential public benefits—enhanced safety assurance and public trust—against increased administrative workload, institutional capacity requirements, and broader economic implications for AV development activity in the state.

**Supporting Institutional Readiness**

Beyond authorization mechanisms, expanded ADS oversight would require supporting institutional capabilities across several areas. Planning functions may need to incorporate AV accommodation into infrastructure investment decisions, including work zone protocols for machine-readable layouts, more current digital representations of lane markings and signal phasing and timing, and integration with Transportation Systems Management and Operations responsibilities. Freight AVs will affect truck flow patterns and corridor support needs; rural mobility implications are also anticipated.

Technical and policy expertise would be needed in areas such as ADS safety concepts, ODD definition and risk minimization, simulation-based scenario analysis, cybersecurity fundamentals, AV-related legal and liability frameworks, and public communications. Inter-agency coordination would be essential, involving interfaces with NHTSA (vehicle recalls), FHWA (roadway standards and digital infrastructure), the FCC (V2X communications), state public utility commissions, and neighboring states for freight and autonomous truck corridors.

Management of mixed-fleet conditions warrants attention regardless of the regulatory model chosen. As AVs operate alongside human-driven vehicles, states may need to update driver education materials, traffic law, and operational procedures to address AV interaction dynamics—including emergency vehicle encounters, traffic stops involving driverless cars, and the impact of AV behavior on congestion and incident clearance. States may also want to update driver manuals and training materials to reflect mixed-fleet realities.

## CONCLUSIONS

This paper has examined how ADS at SAE Levels 4–5 are authorized and overseen across U.S. states, and compared these practices with emerging international regulatory models. Five findings are particularly noteworthy for state policymakers and transportation agencies.

State governments play a central operational role in ADS governance even without formal certification authority. Across the United States, states regulate operation rather than technology design, using registration, insurance, self-certification statements, reporting obligations, and—in some cases—permit-based authorization. Federal oversight remains focused on vehicle design standards, recalls, and post-market enforcement.

State ADS practices cluster meaningfully into three models—baseline legal compliance, administrative self-certification, and permit-based authorization—with increasing regulatory workload and institutional complexity. North Carolina currently aligns with Group 1. The three-model typology provides a practical lens for evaluating trade-offs between safety assurance, administrative burden, and deployment flexibility.

More elaborate regulatory models involve expanded operational governance, not technical certification. States adopting Group 2 or Group 3 approaches focus on clearer operational rules, formal authorization mechanisms, enhanced incident response, structured data access, and defined conditions for suspension or revocation—measures oriented toward accountability, public safety, and transparency rather than algorithmic validation.

A transition toward more elaborate oversight would have significant institutional and planning implications, including the need for workforce expertise in ADS safety and cybersecurity, strengthened inter-agency coordination, public communication readiness, and integration of ADS considerations into infrastructure planning and operations.

International experience offers instructive context, though direct transferability is limited by the U.S. federal–state division of authority. Global convergence toward scenario-based testing, ODD-specific authorization, safety-case requirements, cybersecurity assurance, and lifecycle oversight points to principles that may ultimately shape future U.S. federal frameworks within which states will continue to exercise operational governance.

The proposed Group 1–Group 3 framework provides a practical basis for comparing state AV governance strategies and may assist policymakers in evaluating future regulatory pathways as automated driving systems continue to evolve within the existing U.S. federal–state regulatory structure.

However, the proposed typology is based on comparative analysis of publicly available statutes, regulations, agency guidance, and supporting interviews, and therefore reflects regulatory conditions during the study period. State AV legislation continues to evolve, and individual jurisdictions may transition between regulatory groups over time.

## ACKNOWLEDGEMENTS

This study was conducted as part of a research project sponsored by the North Carolina Department of Transportation (NCDOT). The authors gratefully acknowledge Ms. Sarah Searcy of NCDOT for her guidance and review throughout the project.

## AUTHOR CONTRIBUTIONS

The authors confirm contribution to the paper as follows: study conception and design: George List, Danjue Chen; draft manuscript preparation: Sk Nahia Ahsan, Md. Abdullah Al Hasan. All authors reviewed the results and approved the final version of the manuscript.